ARTICLE

# Code-switching in text and speech challenges information-theoretic speaker design

Debasmita Bhattacharya[1] and Marten van Schijndel[2]

[1]Columbia University in the City of New York, USA and [2]Cornell University, USA
**Corresponding author:** Debasmita Bhattacharya; Email: db3526@columbia.edu

## Abstract

In this work, we use language modeling to investigate the factors that influence insertional code-switching. Code-switching occurs when a speaker alternates between one language variety (the primary language) and another (the secondary language), and is widely observed in multilingual contexts. Recent work has shown that code-switching is often correlated with areas of low predictability in the primary language, but it is unclear whether low primary language predictability only makes the secondary language relatively easier to produce at code-switching points – that is, purely speaker-driven code-switching – or whether code-switching is additionally used by speakers for other purposes, for instance to signal the need for greater attention on the part of listeners. In this paper, we use bilingual Chinese–English online forum posts and transcripts of spontaneous Chinese–English speech to replicate prior findings that low primary language (Chinese) predictability is correlated with insertional switches to the secondary language (English). We then demonstrate that the predictability of the English productions is even lower than that of meaning-equivalent Chinese alternatives, and these are therefore not easier to produce, rejecting the purely speaker-driven theory of code-switching in both writing and speech.

## 1. Introduction

Most of the world's population speaks more than one language (Grosjean & Li, 2013). Among such multilingual speakers, code-switching – the alternation between one language variety and another – is a commonly observed phenomenon (Poplack, 1980). Code-switches are produced by speakers of varying ages (Meisel, 1994) between many different language pairs (Muysken, 2000). For example, below, the Chinese word for *basement*, 地下室, has been replaced by its English equivalent, demonstrating an insertional code-switch:

| | | | | |
|---|---|---|---|---|
| Utterance | 洗衣 | 房 | 在 | basement。 |
| Gloss | laundry | room | in | basement. |
| Translation | *The laundry room is in the basement.* | | | |

Many psycholinguistic and sociolinguistic factors affect code-switching prevalence in bilingual speech. At the utterance level, these include speaker competency, the linguistic context, the speaker's emotional state or affective intention, and the type of information that the speaker wishes to convey, among others (Acuña Ferreira, 2017; Bhattacharya et al., 2024, 2025; Broersma, 2009; Gardner-Chloros, 2009). In contrast to such speaker-centric factors, previous work has proposed a few audience-driven sociolinguistic influences on code-switching, for example, whether the audience is likely to be proficient in all of the relevant languages (Bell, 1984; Dahl, 2009). At the word level, *information load* or *predictability*, as measured by sentence continuation probabilities, has previously been correlated with code-switching (Calvillo et al., 2020; Myslín & Levy, 2015), but as Myslín and Levy point out, this influence could be produced by either speaker-driven or audience-driven pressures.

Since Shannon (1948), information theory has been a prominent framework for understanding communication, in which speakers maximize the communicative function while minimizing the degree to which they overload listeners; this creates a trade-off in optimizing overall communication. As a result, speakers tend to produce predictable and mutually accessible utterances to mitigate the load associated with less predictable expressions, such as jargon. In this work, we investigate whether the relationship between continuation predictability and insertional code-switching is entirely speaker-driven. Under the *speaker-driven* view, which many have assumed (Calvillo et al., 2020; Myslín & Levy, 2015; Valdés Kroff & Dussias, 2023), one (primary) language is generally most salient to a bilingual speaker in a given context. However, in areas of low predictability, which require greater cognitive effort to process, the relative linguistic complexity of the primary language continuation may make the secondary language easier to access for the speaker, increasing the chance that the speaker will switch into the secondary language and produce a code-switch (Beatty-Martínez et al., 2020; Owens, 2005; Poplack, 1980). Alternatively, the same phenomenon could be motivated by other considerations. For instance, an area of low predictability is precisely the place where a listener might get confused. So, a potentially audience-driven speaker, or simply a speaker who is not solely driven by production ease, could signal to the listener that the upcoming material will be challenging and that they should accordingly pre-allocate more cognitive resources (e.g., attention) to the comprehension task to avoid misunderstanding. We speculate that code-switching is a possible mechanism by which the speaker could provide this signal to listeners.

In contrast to prior work that has analyzed the influence of the primary language on the code-switch, we analyze the influence of both the primary and the secondary language on code-switching in both written and spoken communication, and we apply an information-theoretic lens toward investigating whether the role of predictability in shaping code-switching stems purely from speaker-driven pressures. We conclude that speaker-driven pressures do not fully explain the relationship between predictability and insertional code-switching. Our findings support prior results by showing a clear effect of predictability on code-switching, but whereas previous predictability research presumed that this influence was driven by production pressures, our work reveals aspects that cannot be fully explained by production considerations alone. Our research design cannot definitively prove that these results stem from audience design, but we think this would be a fruitful avenue for future inquiry.

## 2. Related work

A few prior studies have challenged the purely speaker-driven theory of code-switching. Sociolinguistic studies like Bell (1984) and Liu (2021) have shown that speakers alter their code-switching behavior based on the features of the listener, providing evidence for identity-based audience design in code-switching. Liu (2021) studied *written* code-switches on the social media platform WeChat. However, this involved in-depth ethnographic analyses of only three users, rather than analysis of broader population-level code-switching behavior. Other work has used eye-tracking to show that readers do use the existence of a code-switch as a cue that foreshadows unexpected information and thereby aids the prediction of subsequent low-probability information (Tomić & Valdés Kroff, 2022). In the *spoken* domain, few studies have considered spontaneous code-switched speech directly through the framework of information theory, though a few (e.g., Fricke et al., 2016; Shen et al., 2020) have shown that bilinguals reduce their speaking rate and modulate low-level phonetic cues prior to a code-switch. Listeners can then attend to these changes and adapt their comprehension of the impending code-switch accordingly. For both spoken and written code-switches, it remains unclear whether producers switch to provide a signal to the comprehender or whether they themselves find it easier to switch at that point for speaker-driven reasons and the resulting cue is incidentally used by comprehenders (Valdés Kroff & Dussias, 2023).

Most similar to our work is that of Calvillo et al. (2020), who studied the influence of predictability on written code-switching, and Myslín and Levy (2015), who studied the same in informal speech. Both Calvillo et al. (2020) and Myslín and Levy (2015) found similar effects and showed that in addition to identity factors, code-switching is correlated with predictability, but again there is the question of *why* this correlation occurs.

From a speaker's perspective, low-probability continuations could simply be less accessible in one language than the other, resulting in a code-switch (Kroll et al., 2006; Sarkis & Montag, 2021). However, multiple studies in the broader psycholinguistic literature suggest that speakers (and writers) adapt their productions to their audience (*audience design*; Clark & Murphy, 1982). For example, in areas of low predictability, speakers produce more function words (Jaeger, 2010; Levy & Jaeger, 2006), and when listeners are engaging in a task that speakers know to be challenging, speakers use more redundant descriptions to communicate with them (Vogels et al., 2019). These findings suggest that there are mechanisms by which speakers can adapt their productions to the needs of listeners, and in this paper, we investigate whether the correlation between predictability and code-switching could be another example of this. Note that our main focus is on interrogating speaker-driven effects, and our foray into audience design is primarily speculative, as prior work has shown that such effects are highly context-dependent, with substantial thresholds for triggering audience adaptations.

Predictability is usually measured by *surprisal*, that is, the negative log probability of a word given its context (Hale, 2001; Levy, 2008; Shannon, 1948):

$$\text{surprisal}(w_i) = -\log_2 \text{P}(w_i | w_{i-t}, \ldots, w_{i-1})$$
$$t = \text{length of preceding context in words}$$

Accurate estimation of this measure depends on large corpora and computational models. Given the relative infrequency of code-switching, it is difficult to capture generalizable code-switching statistics using a standard language model, as restricting the training data to only sentences containing code-switches would require a prohibitive amount of data for training. As a result, Myslín and Levy (2015) asked participants to generate possible continuations for individual contexts but limited the scale of their study to only five speakers. In contrast, Calvillo et al. (2020) used monolingual language models to estimate the above probabilities and were therefore able to perform their analyses over a larger corpus of written online interactions between Chinese–English bilinguals who live and study in the United States (USA). In this paper, we extend the methodology of Calvillo et al. (2020) and use language models to estimate the relative probability of continuations in either language for both written and spoken communication.

## 3. Research questions and hypotheses

While previous work has either focused exclusively on the primary language, examined the characteristics of regions preceding or following a code-switch, or considered only task-oriented settings, we investigate the comprehensiveness of speaker-driven code-switching by studying the features of both the primary language and the secondary language at the point of the code-switch itself (the first word of the code-switched region) in undirected communication settings. By using language models to probe the surprisal of the sentence or utterance in the two different languages, we address our main research questions:

**RQ1:** While there are clearly speaker-driven components to code-switching, does the influence of predictability on code-switching reflect only speaker-centric pressures? More specifically, how does the surprisal of code-switches in the secondary language compare to that in the primary language?

**RQ2:** Does the influence of predictability on code-switching generalize across both writing and speech? In other words, is the relative surprisal of code-switches between the primary and secondary languages consistent in both modalities of language production?

With respect to RQ1, based on previous literature, we expect code-switched regions of writing and speech in the primary language (the language used pre-code-switch) to have low predictability, but speaker-driven and alternative theories, such as audience design, make different predictions about the relative predictability of continuations in each language in the code-switched region. If the comprehensive version of the speaker-driven hypothesis is correct, the predictability of code-switches in the secondary language (the language that is switched into) must be higher than that in the primary language, since the greater accessibility is what triggered the code-switch. Alternative hypotheses permit different information patterns. That is, the unpredictability of some part of the message might induce speakers to produce a code-switch to signal to listeners that a region is difficult to process, even though doing so actually results in a slightly lower predictability than if they had remained in the primary language.

For RQ2, since communication is about transmitting information, we expect information-theoretic constraints, including any influence of predictability, to replicate across written and spoken communication. However, this may occur to differing degrees between modalities as the salience of information-theoretic influences on production could vary by modality, for instance, because only written language allows the producer to revise before committing to a production.

We organize the remainder of the paper as follows: Section 4 describes the two corpora that we augment and use throughout the study. Section 5 introduces the computational methodology and metrics used to process and segment the data. Section 6.1 first aims to verify that written code-switches occur in regions of high surprisal; we compare surprisal distributions from various regions of code-switched and control sentences. Next, since surprisal does not determine predictability in a vacuum, we study how low predictability at code-switches manifests in additional word-level features of writing. These analyses provide essential validation of our written data and corresponding models prior to addressing RQ1 and yield a baseline surprisal of code-switched material, similar to prior studies in this space. We then transition to exploring whether code-switching in writing can be fully accounted for by speaker-driven influences in Section 6.2; we demonstrate this is not the case. Code-switched material exhibits relatively lower accessibility (specifically in terms of predictability) and lower ease of production of the features of code-switches in the secondary language relative to those of its monolingual language patterns. Section 6.3 mirrors Section 6.1, showing that low predictability contributes to spoken code-switching in a preliminary set of analyses, which subsequently address RQ1 for speech and enable cross-modality comparisons for RQ2. Section 6.4 mirrors Section 6.2 by providing evidence against the comprehensiveness of the speaker-driven account of code-switching through the features of the secondary language in speech, answering RQ1 and RQ2. We provide a general discussion of our results and overall conclusions in Section 7.

## 4. Corpora

### 4.1. Written code-switched data

In our work, we first expand the original Calvillo et al. (2020) corpus,[1] comprised of Chinese–English code-switched sentences and their corresponding fully Chinese human translations, which were generated and validated by native Chinese speakers at the time of the dataset's creation to enable analysis with a monolingual Chinese language model. The original Calvillo et al. (2020) corpus also contained a separate set of fully Chinese, non-code-switched, control sentences that lacked code-switches but exhibited syntactic configurations similar to the code-switched sentences. These minimal pairings between sentences that manifested code-switches and those that did not were determined by Calvillo et al. in order to isolate predictability from other factors. The original corpus was collected from web-based, typed, informal communication on public forums between Chinese–English bilingual university students in the USA. The vast majority of the code-switches in the corpus are *insertional* in nature, that is, they involve the insertion of a single word or short phrase in English into an otherwise Chinese sentence. Other types of code-switching, for example alternational switches and congruent lexicalization (Muysken, 2000), are possible, though not present in this dataset. The mean and median lengths of a code-switched region in this corpus are 2.06 and 2.00 words, respectively (standard deviation = 1.39 words). The Chinese portions of the data are written in simplified Chinese. We assume that speakers within the corpus are highly proficient in both languages, given that they are conversing on a Chinese language forum and are attending American universities that have English language proficiency requirements.

[1] https://github.com/lfang1/CodeSwitchingResearch.

To this written corpus, we add automatic English translations of the code-switched sentences and non-code-switched sentences by using the Chinese–English translator Fanyi Youdao. We also experimented with using Google Translate, but that produced notably less fluent and more surprising English continuations; please see Supplementary Appendix B for further comparison between translations produced by Fanyi Youdao and Google Translate. We ensure that the fully English translations retain the original code-switched word through manual examination of the pre- and post-translated data, with changes in inflection when required to maintain correctness in English. The fluency of the translated-to-English sentences is confirmed by two native English speakers. Similarly, the accuracy and meaning preservation of a 10% sample (148 instances) of the automatic translations to English are checked and confirmed by a fluent bilingual Chinese–English speaker. In sum, our expanded version of Calvillo et al.'s corpus contains 1476 originally code-switched sentences, each present twice: once fully translated into Chinese and once fully translated into English. It also contains 1476 non-code-switched control sentences, each present twice: once fully in Chinese and once fully translated into English.

### *4.2. Spoken code-switched data*

We also use a subset of the South East Asia Mandarin-English (SEAME) speech corpus of informal conversations and interviews (Lyu et al., 2010). This corpus originally contained 192 hours of speech corresponding to 256 dialogues between 156 speakers. We filter it to include speech transcripts of utterances that are either code-switched from Chinese to English or fully in Chinese, to maintain consistency with the distinction made in the written dataset between code-switched and non-code-switched sentences. Again, the spoken corpus primarily consists of insertional code-switches. The mean and median lengths of a code-switched region in this corpus are 3.99 and 3.00 words, respectively (standard deviation = 4.19 words). For consistency with the written corpus, we add automatic Chinese translations of the code-switched utterances and automatic English translations of both the code-switched utterances and the fully Chinese non-code-switched utterances. The code-switched subset of the spoken code-switched corpus we use contains 6171 utterances, while the non-code-switched, fully Chinese subset contains 22770 utterances. We use a random sample of 6171 monolingual utterances when performing logistic regressions in Section 6. From a practical standpoint, the scale of our dataset makes human translations intractable and automatic ones a reasonable alternative.

With respect to automatic (machine) translation to English, machine translation is considered a 'mature technology' when one of the languages of interest is English (Isbister et al., 2021), as in our work. This is largely due to the existence of larger, high-performance English language models compared to smaller models in relatively lower-resourced languages. Therefore, we consider it reliable to use automatic translation to English to generate our monolingual English data, as has been done in recent related work such as Iakovenko and Hain (2024).

Further, while we rely on automatic translation to construct some of our monolingual data, we validated our use of this approach by asking two Chinese–English bilinguals to compare automatic and human monolingual translations of a 10% sample (148 instances) of the written code-switched data. They were asked to evaluate which of the two translations they preferred, based on fluency and faithfulness. This validation step revealed that automatic translations were comparable to human translations roughly 50% of the time (73 instances), and, roughly 30% of the time (45 instances), automatic translations were preferred over the human translations, totaling an 80% acceptability rate of the automatic translations (118 instances). The inter-annotator agreement was moderate with Cohen's $\kappa = 0.466$ and, overall, justifies our application of automatic translation to Chinese on the larger spoken code-switched corpus.

Note that both of our corpora consist of *informal* communication, that is, we effectively control for linguistic register across speakers within each dataset and across the two datasets. Topics of informal discussion across both datasets generally overlap (e.g., personal life, academics, professional concerns), enabling an additional coarse-grained control at the domain level.

Our data, analysis code and computational models can be found here: https://github.com/db758/code-switching

## 5. Metrics

Since Calvillo et al. (2020) found that 5-gram surprisal was predictive of code-switching in writing, we train two monolingual Kneser–Essen–Ney smoothed 5-gram language models on Chinese[2] and English[3] Wikipedia datasets, consisting of approximately 360 million and 101 million tokens, respectively. The 101 million-token English dataset is a standard training set for these kinds of English language models. There is no Chinese equivalent of this English dataset, as less work has gone into building standard training sets for Chinese, so we built our own Chinese training corpus using the 360 million-token full Wikipedia dump. Its large size ensures it is as robust as possible, though it contains more tokens than the English training set. Training took roughly 3 hours per model on a laptop CPU. Preprocessing is done using WikiExtractor.

We use these 5-gram models to obtain word-level surprisal values for the fully Chinese and fully English sentences in our written corpus. We also obtain word-level surprisal values using gpt2-base-chinese[4] for our analyses of the fully Chinese data and GPT-2 (Radford et al., 2019) for our analyses of the fully English data, both of which are large language models (LLMs) trained on broad Internet text without reinforcement learning from human feedback. We interface with these LLMs using the minicons Python library (Misra, 2022) and implement correction when computing pseudo-probabilities for words composed of multiple tokens. Throughout this work, we find the same results for LLMs and 5-gram models, so we only discuss the results for 5-grams in the paper. See Supplementary Appendix B for LLM results.

[2]https://dumps.wikimedia.org/zhwiki/latest/zhwiki-latest-pages-articles.xml.bz2.
[3]https://huggingface.co/datasets/Salesforce/wikitext.
[4]https://huggingface.co/ckiplab/gpt2-base-chinese.

To obtain word-level surprisal values for our spoken data, we train two 5-gram language models on monolingual conversational speech transcripts from the CallHome (English; Canavan et al., 1997) and HUB5 (Chinese; Linguistic Data Consortium, 2018) corpora. These consist of approximately 170 and 190 thousand tokens, respectively. We note that these models are several orders of magnitude smaller than those we trained on written text. While future work could train larger models and attempt to replicate our spoken analysis to confirm the reliability of our findings, we point out the importance of using smaller, and therefore more accessible and less resource-intensive, models.

For our analyses on written code-switches, we partition our data into different types. We distinguish between sentences that originally contained code-switches (**CS**) and fully Chinese control sentences that lacked any code-switch (**Non-CS**). Since our method directly uses the fully Chinese translations of the original code-switched sentences, instead of the original Chinese–English code-switched sentences themselves, we also refer to these translated Chinese outputs as **CS** sentences and differentiate them from the control Non-CS Chinese sentences that never contained code-switches. Calvillo et al. paired each original CS sentence with a control Non-CS sentence that has a Chinese word in a similar syntactic configuration to that of the first code-switched, that is, English, word in the original CS sentence.[5] We designate the word at the site of this 'first code-switched word' in both CS and corresponding Non-CS sentences as **CS1**. Note that CS1, that is, the word at the code-switch site, surfaces in English only in original CS sentences; in the translated CS sentences, CS1 is the Chinese equivalent of the original English code-switched word, and in Non-CS sentences, CS1 is simply a Chinese word whose location and syntax match that of the English code-switched word in the corresponding original CS sentence. We designate all other words in Non-CS sentences, that is, all words that are not CS1, as **Non-CS words**. See Table 1 for a visual schematic of these regions. We determine Chinese word segmentations using the Jieba Python library's tokenizer.

**Table 1.** Schematic illustrating the shorthand terminology used throughout the remainder of this paper on (1) original code-switched sentences, (2) code-switched sentences that have been fully translated to Chinese and (3) control non-code-switched sentences in the written dataset.

| Sentence type | Sentence sub-type | Example | Notes |
|---|---|---|---|
| (1) CS [orig.] | – | 你 是 full-time 学生 吗? | CS1 |
| | Gloss | You are full-time student *part.*? | |
| | Translation | *Are you a full-time student?* | |
| (2) CS [trans.] | – | 你 是 全日制 学生 吗? | |
| | Gloss | You are full-time student *part.*? | |
| | Translation | *Are you a full-time student?* | |
| (3) Non-CS | – | 是 里程 票 吗? | CS1 |
| | Gloss | Is mileage ticket *part.*? | Non-CS |
| | Translation | *Is it a mileage ticket?* | |

[5]Further information about how 'syntactically similar configurations' were defined and how Non-CS sentences were identified in the original Calvillo et al. corpus is available in Calvillo et al. (2020). In the current work, we simply make use of this pre-determined set of control Non-CS sentences.

For our analyses on spoken code-switches, we similarly distinguish utterances that originally contained code-switches (**CS**) from fully Chinese utterances (**Non-CS**). However, only the **CS** utterances in the spoken code-switched corpus contain a **CS1** word, as the **Non-CS** utterances are not syntactically paired with **CS** utterances in this dataset, that is, all words in **Non-CS** utterances are **Non-CS words** in the spoken corpus of code-switches.

As is the case with most open-sourced code-switched datasets, speaker metadata on language proficiency is not provided in either of our corpora. Therefore, we focus on word-level surprisal as our primary speaker-centric variable of interest and exclude from the study alternatives such as differences in cognitive processing or information load tolerance, for which we lack data to appropriately operationalize.

## 6. Experiments

### *6.1. Low predictability contributes to code-switching at CS1 in writing*

We begin by using our Chinese language model to confirm the Calvillo et al. (2020) finding that code-switches occur in regions of high surprisal in writing. That is, we compare the distribution of Chinese CS1 surprisal between CS and Non-CS sentences (Figure 1). We confirm that CS1 is significantly more surprising in CS sentences than in Non-CS sentences ($p$ = 1.448e−09), but the surprisal difference between the sentence types is very small, with a mean Δ of 1 bit. Note that throughout this paper, we determine significance using Welch's two-sample t-tests in R 3.6.2.

However, the CS1 surprisal values in both CS and Non-CS sentences are significantly higher (both $p$ < 2.200e−16) than the surprisal values of Non-CS words in Non-CS sentences. We also find that the surprisal of Non-CS words is similar to that of held-out monolingual Chinese Wikipedia data (Supplementary Figure B1 in Supplementary Appendix B). To summarize, mean surprisal in fully Chinese control sentences that have never been code-switched is much lower than at CS1 (mean Δ

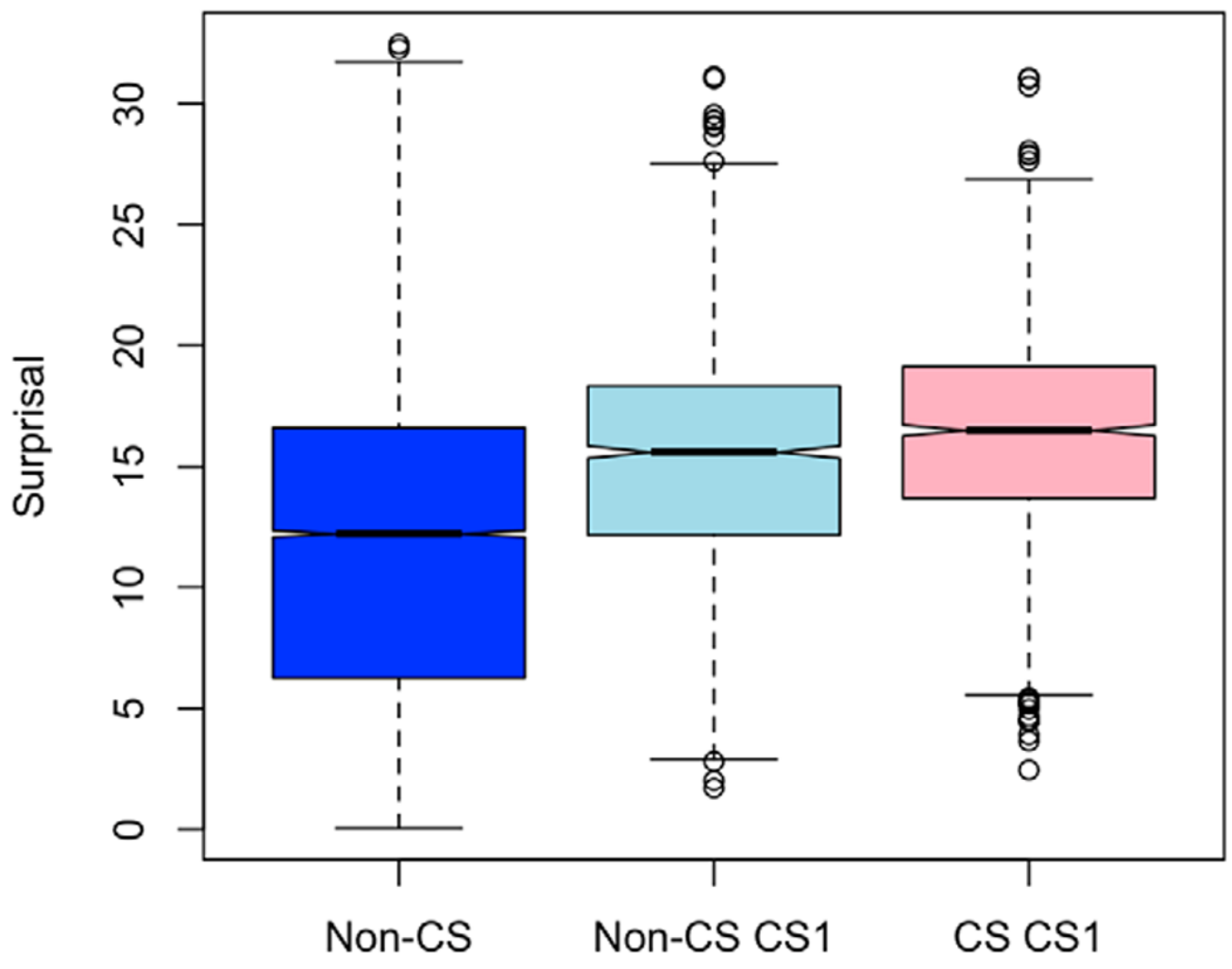

**Figure 1.** Comparing surprisal of CS1 words in code-switched (CS) and non-code-switched (Non-CS) sentences to Non-CS words in Non-CS sentences. Boxplot whiskers extend to 1.5x the interquartile range.

from CS1 is 4.39 bits and 3.41 bits in CS sentences and Non-CS sentences, respectively).

Overall, these results support the conclusions of Calvillo et al. (2020) that code-switched words occur in regions of low predictability in writing; however, the surprisal of CS1 words in non-code-switched sentences is generally much higher than that of Non-CS words, suggesting that a word at the CS1 site, even in cases where no code-switching occurs, differs from standard monolingual written Chinese. We replicate this finding using alternative surprisal values of CS1 in CS sentences from additional manual human translations of English words at CS1 to Chinese (Supplementary Appendix A).

Given the non-deterministic nature of code-switching, we interpret this as an indication that CS1 sites are generally regions of low predictability that are likely to exhibit code-switching (as occurred in the paired CS sentences in the original version of the written corpus). By comparing CS1 in CS sentences with CS1 in Non-CS sentences, Calvillo et al. (2020) demonstrated that surprisal is a driver of code-switching after controlling for the other influences that define CS1 sites. However, ideally, we would like to quantify the influence of any factors that differentiate CS1 sites from monolingual Chinese, rather than only those factors that push CS1 sites to switch more often.

To better quantify the full impact of surprisal alongside other factors that influence written code-switching, we implement a binary logistic regression model using scikit-learn 1.1.0 to differentiate between CS1 words in CS sentences and Non-CS words in Non-CS sentences. For the CS sentences, we use the characteristics of CS1. For the Non-CS sentences, we randomly sample a Non-CS word rather than CS1 itself, unlike Calvillo et al.

We use most of the same regression features as those used by Calvillo et al. We partition part-of-speech tags into three categories: nouns, verbs, and other. We use dummy-coding with *noun* as the reference tag (i.e., all part-of-speech results are relative to nouns). We replace sentence length with the relevant word's 0-indexed position in the sentence. We find in our data that code-switched sentences tend to be longer than non-code-switched ones (Supplementary Figure B2 in Supplementary Appendix B). Given this, we are interested both in finding out whether code-switches occur later in these longer sentences, and in keeping all of the features of our regression consistent with *word-level* analyses. We also replace word length in terms of Chinese character length with word length in terms of pinyin character length, a metric that more directly corresponds to typing effort via number of keystrokes.[6] We obtain these values by using the python-pinyin library, accessed via https://github.com/mozillazg/python-pinyin/tree/master. We standardize all numerical features – that is, word length, word frequency, word position and surprisal – used in the regressions such that their means are 0 and their standard deviations are 0.5, in keeping with the methodology followed by Calvillo et al. We transform frequency values with a negative $\log_2$ operation to make the values comparable to our surprisal values, meaning that a large frequency value corresponds to a less common word in our analyses (i.e., frequency is 'unigram surprisal').

[6]Due to the logographic nature of Chinese orthography, there is very little variance in word lengths in terms of Chinese character length in our written corpus. Without conversion to pinyin word length, median word length is greater for CS1 words than for Non-CS words, but Non-CS words are some of the longest words in the written dataset (see Supplementary Table B1 in Supplementary Appendix B). Mean word length in terms of Chinese character length is fewer than two characters, regardless of region.

**Table 2.** Summary of the logistic regression model ($R^2$ = 0.862) for CS1 (coded 1) versus random Non-CS1 (coded 0) in the dataset of written code-switches.

| Factor | coef | std err | *t* |
|---|---|---|---|
| Intercept | 0.5113 | 0.006 | 88.964 |
| POS = verb | 0.0162 | 0.010 | 1.608 |
| **POS = other** | **−0.0371** | 0.009 | **−3.910** |
| **Frequency** | **0.8884** | 0.008 | **118.065** |
| **Word length** | **0.0578** | 0.009 | **6.390** |
| **Word position** | **−0.0184** | 0.007 | **−2.642** |
| **Surprisal** | **0.0199** | 0.009 | **2.270** |

*Note:* Bolded features are significant.

Our regression supports our previous finding that higher surprisal is correlated with code-switching (Table 2). In line with prior studies, we also find that rarer Chinese words are more likely to be code-switched (Calvillo et al., 2020; Forster & Chambers, 1973), longer Chinese words, that is, words that require greater typing effort, are more likely to be code-switched (Calvillo et al., 2020; Myslín & Levy, 2015), and non-noun/non-verb parts of speech are less likely to be code-switched (Myers-Scotton, 1993; Myslín & Levy, 2015). We unexpectedly find that later word positions in a sentence are less likely to be code-switched, which contrasts with our and Calvillo et al.'s observation that longer sentences, which offer greater scope for later word positions than do shorter sentences, tend to contain code-switching. Overall, our results confirm that higher surprisal regions are more likely to exhibit code-switching in writing. We replicate these results with surprisal values from an LLM, in this case, gpt2-base-chinese (Supplementary Table B2 in Supplementary Appendix B).

### 6.2. Code-switching in writing has information-theoretic components that are not solely speaker-driven

We next examine the features of the secondary language, English, at CS1 in CS sentences. Recall that we are studying insertional code-switching, where English is inserted into Chinese sentences. Thus, all following mentions of 'code-switched English' refer to code-switches (from Chinese) to English.

To start, we observe that many of the code-switched English words in the corpus are jargon-like (i.e., highly domain-specific) and/or likely to be used far more in secondary language contexts (in this case, English college) than in primary language contexts (in this case, Chinese pre-college). This is expected from an information-theoretic perspective, as one way in which speakers can maximize their communicative efficiency is by using such expressions, which are in turn dependent on different, often speaker community-based, factors. In this corpus, such words include rental-related terms like 'lease', 'apartment', 'garage', 'utility', 'studio' and 'house', as well as terms to do with bed sizes such as 'king' and 'queen', accounting for about 20% of the code-switched words in the corpus (295 instances). Technical terms like 'email' and 'PhD' also appear frequently, accounting for an additional 5% of the code-switched English in the corpus (74 instances), while proper nouns like 'JFK' account for almost 1% of the corpus (13 instances). These observations highlight some of the wide variety of possible sociological factors behind code-switching that are important but that we do not explore further in this work, as we are unable to explicitly identify or control for them in the data. While offering no sociolinguistic claims, we note that since students are less likely to have leased apartments on their own prior to attending university in the USA, the fact that they use English terms for leasing is not surprising, even if English is their less dominant language. However, it is not informative for our purposes to examine cases in which speakers cannot use Chinese and are instead forced to switch to English; what is relevant for our investigation of predictability are cases where they choose to switch even in contexts for which they otherwise could have used Chinese.

In addition to observations on what constitutes viable data for our investigation, since our models for analysis of written code-switches are trained on Wikipedia data that do not often cover such jargon-like topics, we have reason to believe that these models would not account for the sociological factors underlying written code-switching in our data and would not be able to correctly model the jargon-like terms. Testing our models on such terms would then likely produce unreliable measures of surprisal. To avoid this effect and focus only on relevant data, we remove the jargon-like terms outlined above (corresponding to 26% of the corpus) before conducting the remaining analyses in this section. However, we confirm that performing our analyses on the subset of CS1 data that includes the jargon-like terms produces effects with identical signs to our reported results, though the magnitudes of the effects differ (see Supplementary Appendix B for details).

We also consider a potential confound of this study, whereby people may code-switch when there is a more precise way to express their intended meaning in the language they are code-switching to. To address this confound, we asked three native Mandarin speakers to annotate a subset (10% sample; 148 instances) of the data, comparing English and Chinese translations of each code-switched sentence and marking which of the two is more precise in meaning with respect to the target (code-switched) word by coding Chinese as 0, English as 1, and equal precision as 2. We find that in more than two-thirds of the cases, the Chinese continuation is more precise than or equally as precise as the English alternative (Krippendorff's $\alpha = 0.218$, i.e., fair agreement), which suggests that semantic precision is not a dominant factor influencing code-switching in the vast majority of cases. This, in turn, implies that precision of meaning is not a significant confound of the study. Since our data are not significantly confounded by considerations of semantic precision, we proceed to study code-switching purely through a lens of predictability and leave further investigation of additional semantic factors, such as recency of exposure or domain activation, to future work.

#### *6.2.1. CS1 English is more complex than monolingual English in writing*

To determine whether CS1 English words are as accessible as we would expect under a purely speaker-driven framework, we perform baseline comparisons to standard written monolingual English. We first compare the lengths of English words at CS1 to the monolingual English present in Wikipedia, which we treat as a proxy for written English. Since information load is high at CS1, we know that this encourages a speaker to code-switch from Chinese to English in our corpus. If they remain in Chinese, the speaker is likely to produce a high surprisal but short Chinese word; the high surprisal reflects what we know about information load at CS1, while the word would be short due to shorter words being easier to produce than longer words, in alignment with intuitions stemming from a purely speaker-driven framework. If the speaker chooses instead to code-switch to English at CS1, the relevant question becomes: What kind of *English* word is produced relative to possible English alternatives?[7] A code-switched English word that is shorter than average would suggest high accessibility and ease of production in alignment with purely speaker-driven theories; a relatively longer code-switched English word at CS1 would imply the opposite. We find that a code-switched English word at CS1 tends to be longer than the average monolingual word in English (Figure 2a; $\Delta$ = 0.47 characters; $p$ = 6.895e-15). This is in contrast with previous speaker-driven work such as Forster and Chambers (1973), which showed that speakers found shorter words more accessible than longer ones.

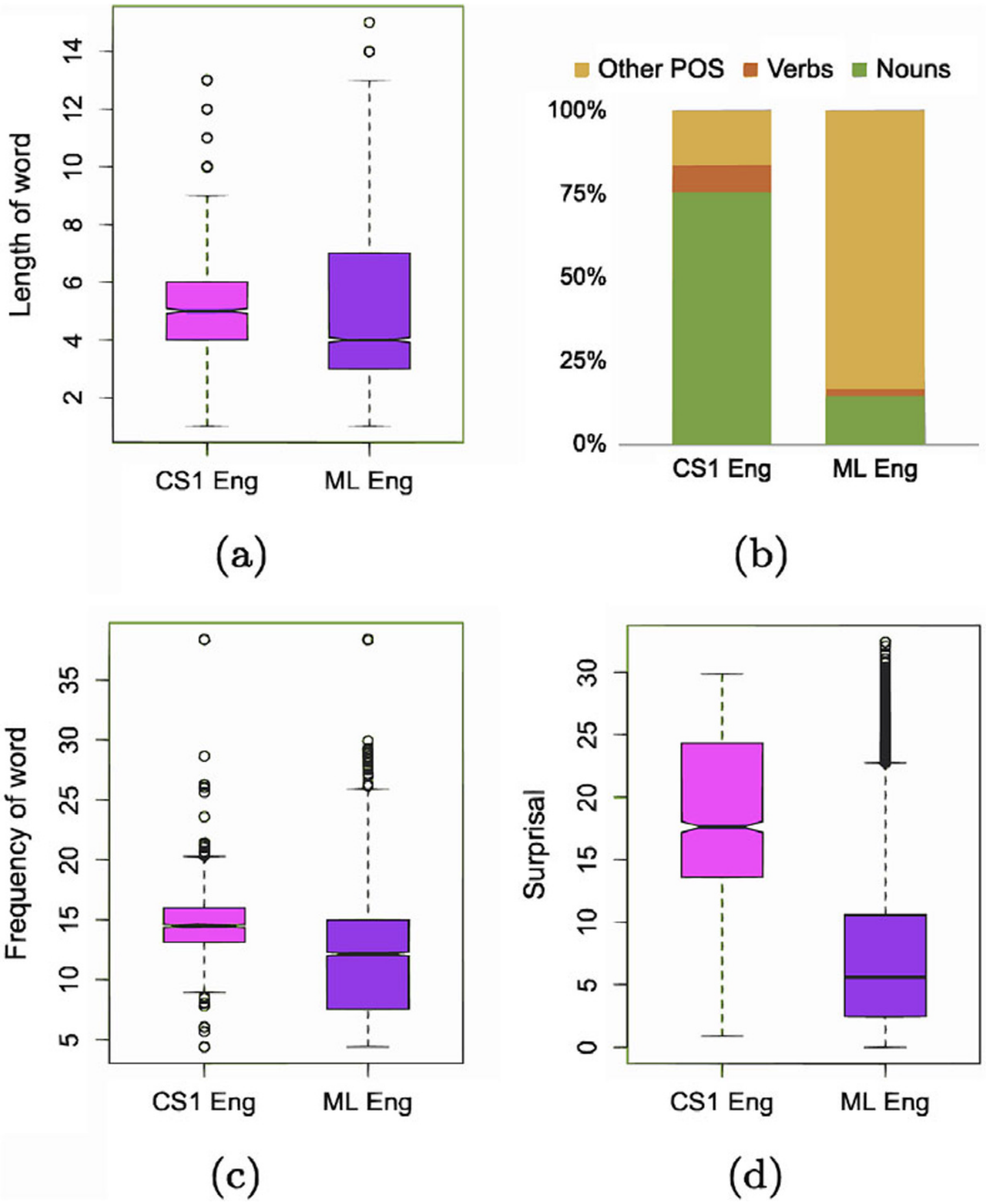

**Figure 2.** Comparing CS1 in English and the monolingual (ML) English vocabulary across (a) word length, (b) part-of-speech tag distribution, (c) word frequency and (d) surprisal, in writing.

[7]In this study, we restrict our cross-linguistic comparisons to questions of predictability in each language. While there are other factors that affect salience, such as word length, it remains beyond the scope of this paper to explore ways of directly comparing these measures cross-linguistically.

Next, we compare the distribution of part-of-speech tags at English CS1 to that in written English. Recall that we have binned parts of speech into *noun*, *verb* and *other*. We find that most English words at CS1 are nouns (Figure 2b). This aligns with the Closed Class Constraint, which suggests that closed-class items, such as pronouns, prepositions, conjunctions, etc., cannot be code-switched except as part of a larger phrase (Joshi, 1982),[8] as well as previous work including Myslín and Levy (2015), which has found that nouns are the most code-switched part-of-speech, followed by verbs and finally all other parts of speech. However, in our written data, verbs are less common than other parts of speech. This may be driven by the fact that this analysis is based on code-switching in writing, which may contain more non-verb modifiers than in the previous speech-based work from the literature. In contrast, we find that most words in monolingual English are non-noun/non-verb parts-of-speech (Figure 2b). The second most represented are nouns, followed by verbs.

Finally, we analyze the word frequencies and surprisal of English CS1 words and monolingual English. CS1 words in English are significantly less frequent than monolingual English words (Figure 2c; $p < 2.200e{-}16$), and similarly, English CS1 words are much more surprising than monolingual English (Figure 2d; $\Delta = 11.23$ bits; $p < 2.200e{-}16$). We replicate these surprisal results with surprisal values from an LLM, in this case, GPT-2 (Supplementary Figure B6 in Supplementary Appendix B). These findings contrast with previous work (e.g., Calvillo et al., 2020; Forster & Chambers, 1973), which speculated that higher frequency and less surprising words may be easier to access, and thus may be more likely candidates for speaker-driven code-switches than lower-frequency and more surprising words. However, since these prior findings were based only on analyses of the primary language (Chinese, in our case), previous studies could not test this hypothesis.

So far, our exploration of the features of the secondary language has suggested that code-switched written English is different from standard written monolingual English along several dimensions. Our investigation of word length, word frequency and surprisal has reinforced this notion by demonstrating that code-switched written English at CS1 tends to be more difficult to process (in the case of word length) and access (in terms of word frequency and surprisal), or, in other words, more linguistically complex than standard written monolingual English.

In addition, our exploration of the features of the secondary language has indicated that the characteristics of CS1 English words are different from those of code-switches examined in previous work. In particular, CS1 English words are less accessible than we would expect them to be under a purely speaker-driven framework. Thus, we have some initial indications of pressures beyond production ease underlying the code-switching occurring in our written data. These findings may be attributed to the fact that our work differs from previous studies in examining the secondary language involved in instances of code-switching, and not just the primary language.

#### *6.2.2. CS1 English is relatively more complex than CS1 Chinese in writing*

Our findings so far suggest that English CS1 code-switches may be relatively cognitively costly to produce in writing. Under a purely speaker-driven account based on production ease, CS1 in Chinese would need to have an even larger complexity difference with standard monolingual Chinese than CS1 in English does with standard monolingual English. Such an overriding complexity difference in Chinese would support the idea that the English continuation is relatively simpler than continuing in Chinese, in effect making it easier for a speaker to switch to producing English in writing than remaining in Chinese.

[8]While a few, more recent studies have documented code-switching patterns that do not fully conform to this constraint, other linguistic theories, for example Prince and Pintzuk (2000), suggest that these instances may in fact constitute speech errors in the form of language interference, rather than true code-switching that would violate Joshi (1982)'s proposed governing constraint.

To make this comparison, we compare the relative surprisal of CS1 words in each language to the mean surprisal that can be expected in the vocabulary of that language (Figure 3). To obtain the relative English surprisal, we normalize each CS1 English surprisal value by the mean surprisal value of the monolingual English Wikipedia data. Similarly, we normalize each Chinese CS1 surprisal value by the mean surprisal value of the monolingual Chinese Wikipedia data.

We find that CS1 words in English are significantly more surprising than in Chinese, even after normalizing by the expected surprisal for the monolingual vocabulary of each language (Figure 3). This is also true for surprisal sourced from LLMs (Supplementary Figure B7 in Supplementary Appendix B). Not only is CS1 in English more complex than monolingual English according to a number of measures, it is even more surprising than the Chinese continuation in the CS sentences would have been in comparison with monolingual Chinese. Thus, code-switching into English at CS1 seems to make a sentence less information-theoretically accessible to the writer than if they had remained in Chinese. This provides further evidence against the comprehensiveness of the speaker-driven predictability hypothesis. That is, because the Chinese word at CS1 is harder than typical Chinese continuations, producers code-switch to English for reasons other than their own ease of production; we speculate that one potential example of such a reason would be to signal that difficulty for their audience. This provides an information-theoretic complement in the written domain to previous findings on spoken audience-driven cues of a similar kind, for example phonetic signals. We next test whether our information-theoretic results on writing generalize to speech as well.

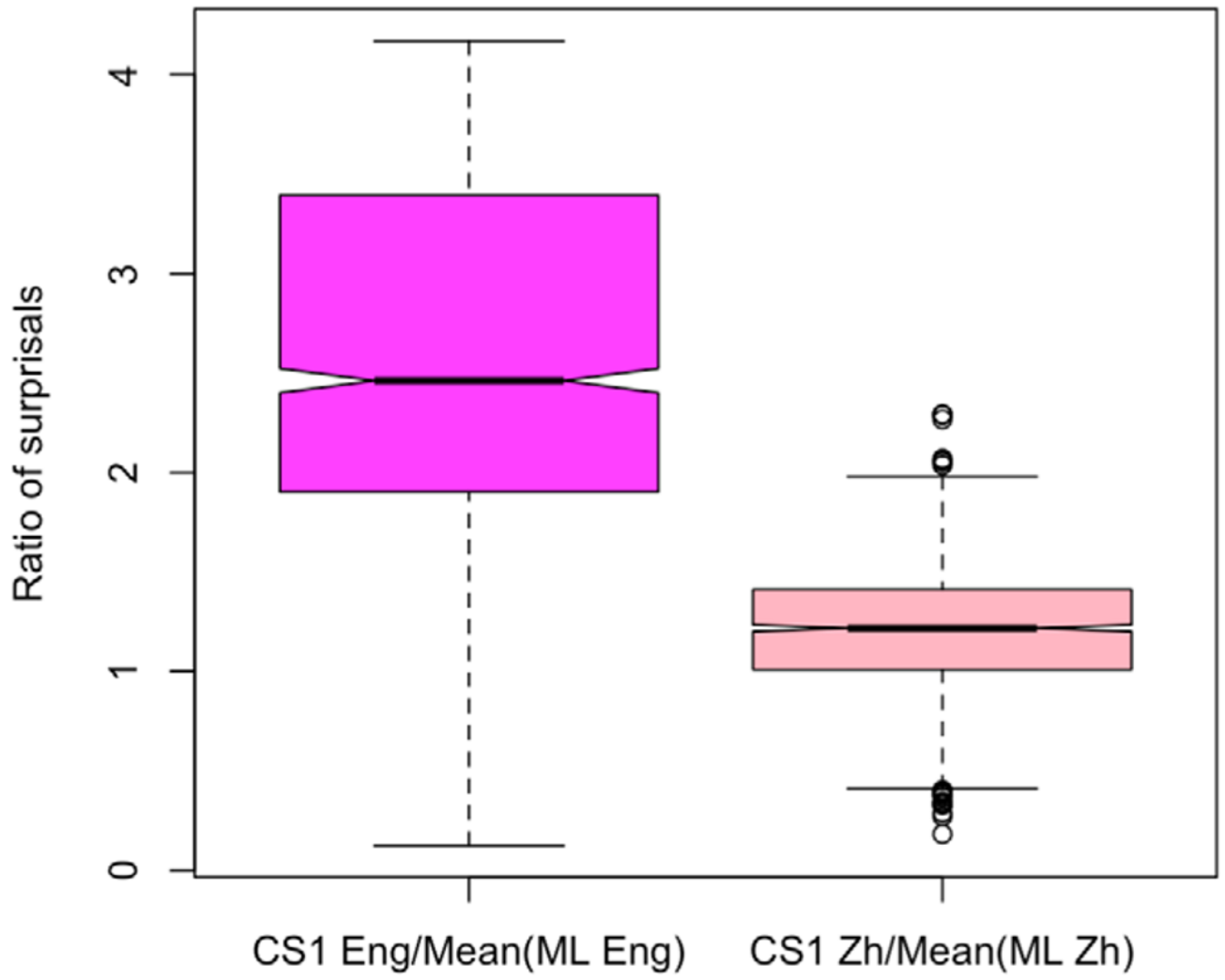

**Figure 3.** Comparing normalized CS1 surprisal in English to Chinese in writing.

### 6.3. Low predictability contributes to code-switching at CS1 in speech

We now examine the role of predictability on spoken code-switching. We use another binary logistic regression to quantify the impact of surprisal on code-switched speech by distinguishing Chinese CS1 words in CS utterances from Non-CS words in Non-CS (i.e., fully Chinese) utterances. Since the Non-CS utterances in our dataset of spoken code-switches do not each contain a CS1, we simply choose a random word in each fully Chinese utterance to represent the characteristics of a Non-CS word. In the aggregate, these 6171 samples provide us with the characteristics of Non-CS language. For the CS utterances, we use the characteristics of the CS1 word, as before.

We use all the same regression features as for our written code-switched data, with the exception of measuring word length in terms of phonemes rather than number of keystrokes, to account for the modality of the data. This is calculated using the phonemizer Python library. Again, our regressions reveal that higher surprisal is correlated with code-switching (Table 3).

Similar to our previous written results, we find that longer Chinese words made up of a greater number of phonemes, earlier word positions in an utterance, and nouns, relative to verbs and other parts of speech, are more likely to be code-switched. Inconsistent with the speaker-driven hypothesis, we also find that more frequent Chinese words are more likely to be code-switched in speech. For the most part, our results on code-switched speech align with those we have previously found on code-switched text, and confirm that higher surprisal regions are more likely to exhibit code-switching in speech.

### 6.4. Code-switching in speech has information-theoretic components that are not solely speaker-driven

#### 6.4.1. CS1 English is more complex than monolingual English in speech

As before, we examine the features of the secondary language, English, at CS1 in code-switched utterances. Unlike our written dataset, our dataset of spoken code-switches spans a much wider variety of topics and thus does not display any obvious groupings of jargon-like words. Given this, we retain all the data for the remainder of our analyses on the spoken code-switches.

Similar to our previous analyses, we first compare the lengths of English words at CS1 to the monolingual English present in the speech transcripts of the CallHome corpus, which we treat as a proxy for spoken English. Again, the code-switched English word at CS1 tends to be longer than the average monolingual English word (Figure 4a; $\Delta = 2.013$ phonemes; $p < 2.200\text{e}{-}16$).

**Table 3.** Summary of the logistic regression model ($R^2 = 0.290$) for CS1 (coded 1) versus random Non-CS1 (coded 0) in the dataset of spoken code-switches.

| Factor | coef | std err | *t* |
|---|---|---|---|
| Intercept | 0.6811 | 0.008 | 86.925 |
| **POS = verb** | **−0.2605** | 0.012 | **−22.422** |
| **POS = other** | **−0.2819** | 0.012 | **−24.224** |
| **Frequency** | **−0.2695** | 0.010 | **−26.817** |
| **Word length** | **0.1771** | 0.010 | **18.049** |
| **Word position** | **−0.0554** | 0.008 | **−7.257** |
| **Surprisal** | **0.3160** | 0.013 | **24.856** |

*Note:* Bolded features are significant.

As before, most English words at CS1 are nouns, followed by other parts of speech and verbs (Figure 4b). Again, most words in monolingual English speech are non-noun/non-verb parts of speech. This is followed by verbs and then nouns.

Finally, we analyze the word frequencies and surprisal of English CS1 words and monolingual English. CS1 words in English are significantly less frequent than monolingual English words (Figure 4c; $p < 2.200e{-}16$), and similarly, English CS1 words are much more surprising than monolingual English (Figure 4d; $\Delta = 6.515$ bits; $p < 2.200e{-}16$).

Our exploration of the features of the secondary language has again suggested that code-switched spoken English is not only different from but also more linguistically complex than standard (conversational) spoken monolingual English.

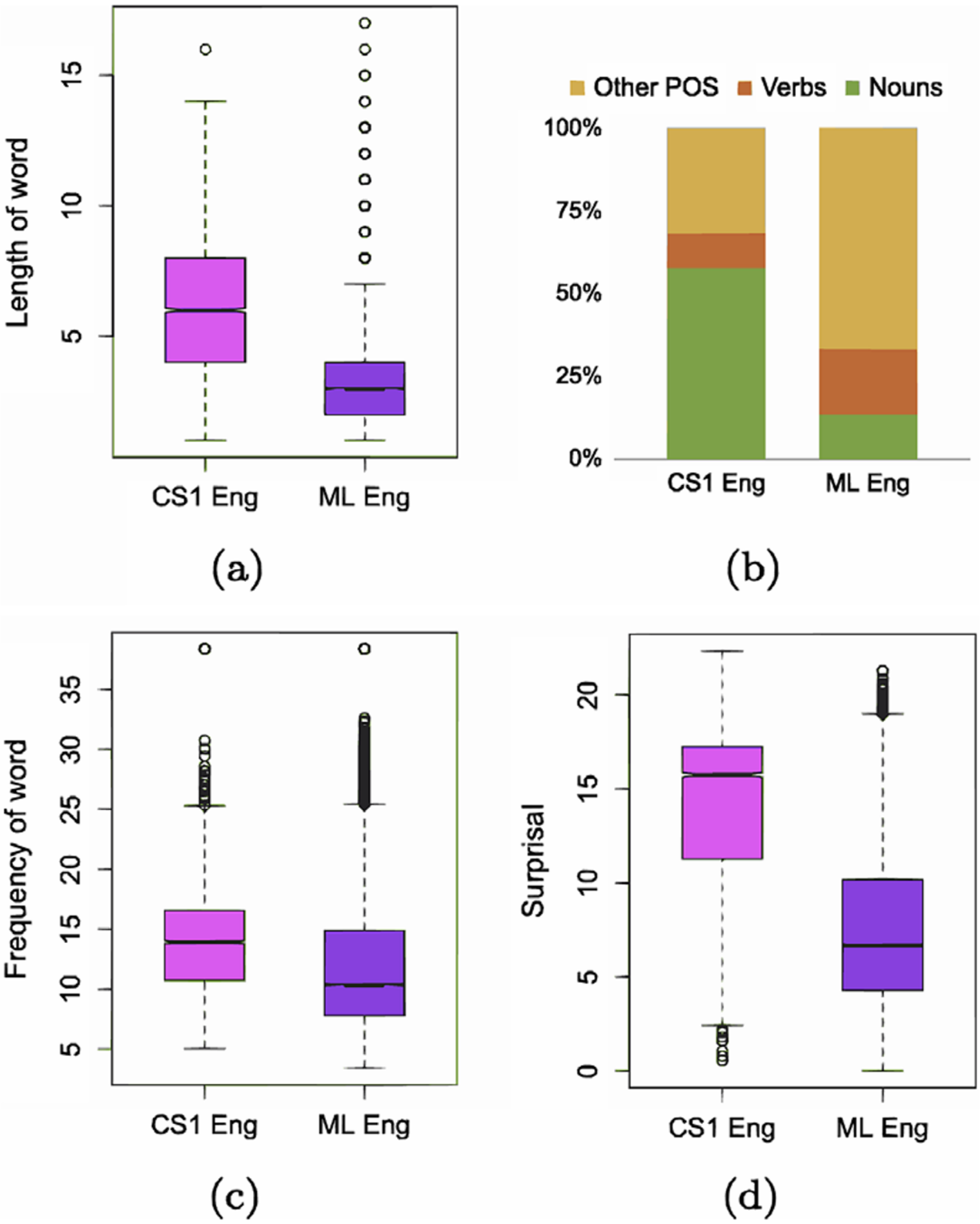

**Figure 4.** Comparing CS1 in English and the monolingual (ML) English vocabulary across (a) word length, (b) part-of-speech tag distribution, (c) word frequency and (d) surprisal, in speech.

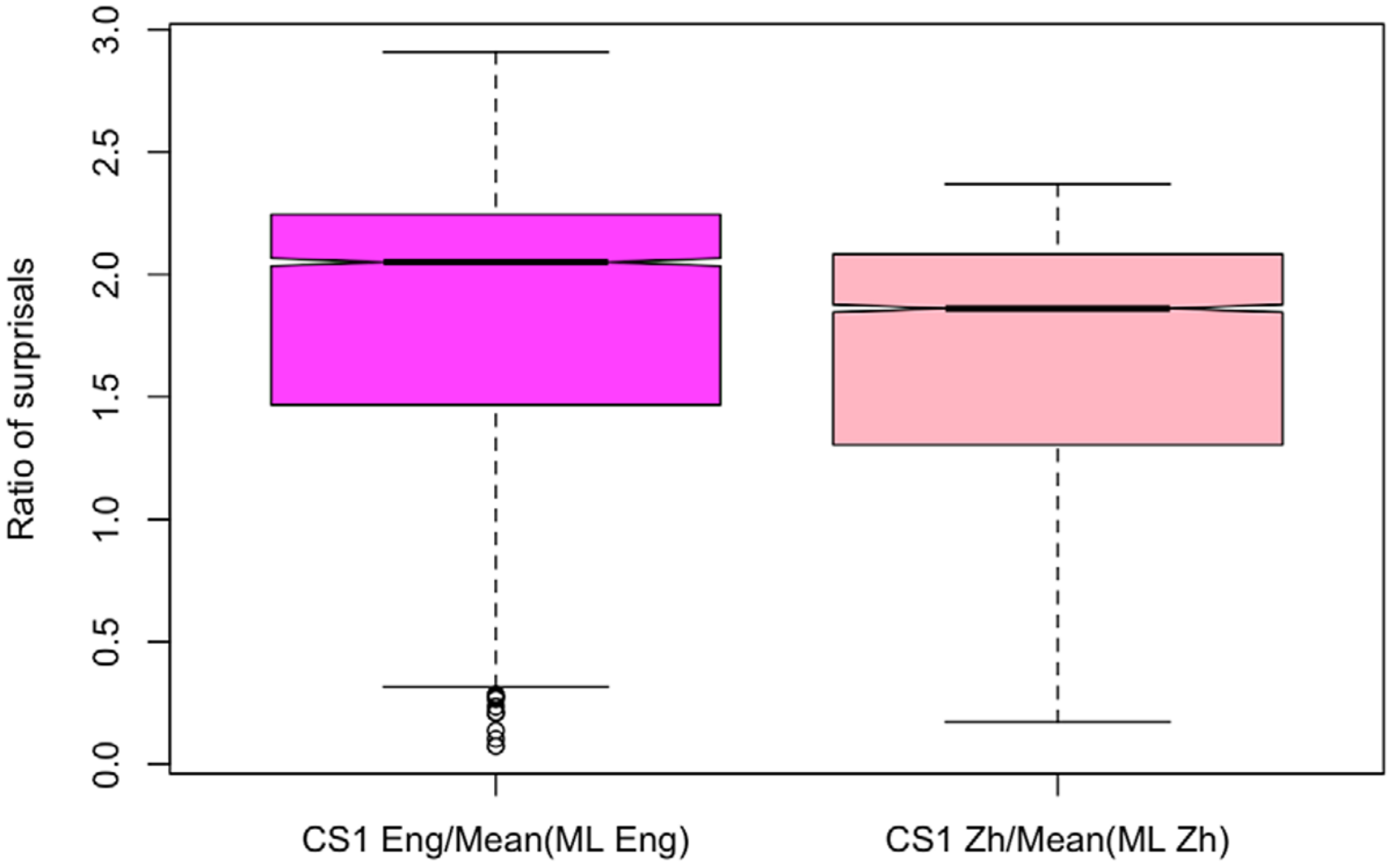

**Figure 5.** Comparing normalized CS1 surprisal in English to Chinese in speech.

#### *6.4.2. CS1 English is relatively more complex than CS1 Chinese in speech*

Finally, we compare the relative surprisal of CS1 words in each language to the mean surprisal that can be expected in monolingual productions of that language (Figure 5). Similar to before, we normalize each CS1 English surprisal value by the mean surprisal value of the monolingual CallHome English data and normalize each Chinese CS1 surprisal value by the mean surprisal of the monolingual HUB5 Chinese data.

We find that CS1 words in English are significantly more surprising than in Chinese, even after normalizing by the expected surprisal for monolingual productions of each language. In other words, CS1 in English is both more complex than monolingual English productions across several linguistic features *and* more surprising than CS1 in Chinese is in comparison with monolingual Chinese productions. Thus, code-switching into English at CS1 makes an utterance even more surprising and thus likely less salient for speakers during production than if the speaker had remained in Chinese. As with our analyses of written code-switches, we have found evidence of pressures on predictability that do not seem to uniquely make production easier for the speaker, and appear to play an additional role in using code-switching to help indicate regions of linguistic complexity for the listener.

## 7. Discussion

In this work, we use language modeling to investigate whether speaker-driven factors can fully account for an information-theoretic influence on Chinese–English code-switching. Rather than focus exclusively on analyses of the primary language (Chinese, in this case) to draw conclusions about code-switching, as has been common in prior work, we conduct multiple analyses of the secondary language (English, in this case) as well, enabling us to resolve a number of details about insertional code-switching and its relationship with predictability in bilingual online forum posts and transcripts of spontaneous speech.

All of our analyses confirm prior findings that predictability influences code-switching (Calvillo et al., 2020; Myslín & Levy, 2015). We additionally find that code-switched English is more linguistically complex than standard monolingual English and that its features make code-switches to English less accessible than one would expect under a speaker-driven framework. Furthermore, while surprisal in the primary language is higher at code-switched regions relative to non-code-switched Chinese, the surprisal in the secondary language is even higher relative to non-code-switched English. These patterns hold true across both written and spoken modalities. Combined, these results suggest that insertional code-switching actually increases the information load compared to a counterfactual monolingual production, though this effect is somewhat more pronounced for written code-switches compared to spoken ones. This provides strong evidence that code-switching in either modality cannot be fully explained by the speaker-driven account that views code-switching as a method for easing production. Our cross-modality finding is striking, given that influences on production may differ between writing and speech; in particular, spoken language does not offer the producer an opportunity to revise, suggesting that our findings result from online production processes rather than subsequent revision considerations. Nonetheless, our results are consistent across modalities of language production, indicating the salience and general uniformity of the effect we uncover. Additionally, we replicate our results from traditional *n*-gram language models using large neural language models and thus show that language models of either kind are appropriate tools for identifying the patterns we find in our work. We speculate that the extent to which these patterns manifest may be tied to the specific language pair under investigation, especially since English and Mandarin are highly typologically distinct languages, with little overlap in terms of cognates, syntax or even writing systems. We anticipate that future studies on additional language pairs, varying in typological distance, will provide further insight into how the specific languages involved in code-switching modulate the effects presented here.

Our work is the first to analyze both primary and secondary language surprisal across written and spoken modalities in the information-theoretic study of code-switching. We extend prior information-theoretic analyses to show that speaker production ease cannot fully account for code-switching patterns. We interpret our findings as indirect evidence of audience design pressures, though further study is required to conclusively prove this connection. Our analyses offer a novel perspective on explaining the motivation underlying code-switching behaviors. This has the potential to open new lines of research on multilingual language production while broadening existing information-theoretic perspectives, which together will help address the fundamental question of *why* writers and speakers code-switch.

**Data availability statement.** The original written data consisting of 3k code-switched and control sentences with corresponding Chinese translations, which we did not collect or generate, can be found at the following link: https://github.com/lfang1/CodeSwitchingResearch. We added 3k English translations to this written dataset. The original spoken data consisting of 101k transcribed utterances, which we did not collect or generate, can be found at the following link: https://catalog.ldc.upenn.edu/LDC2015S04. We used a subset of these 101k utterances: 6k were code-switched, and 23k were not code-switched. We added Chinese and English translations of this subset of utterances, totaling an additional 58k utterances. All of our materials, augmentations to the original datasets, analysis code, and computational models can be found at the following link: https://github.com/db758/code-switching. Throughout the paper, we use Welch's two-sample t-tests to test for statistical significance in independent comparison groups that may have unequal variance and/or sample sizes.

**Competing interests.** The authors declare none.